\documentclass[a4paper,fleqn]{cas-sc}

\usepackage[authoryear,longnamesfirst]{natbib}

\usepackage{algorithm}
\usepackage{algpseudocode}
\usepackage{graphicx}

\def\tsc#1{\csdef{#1}{\textsc{\lowercase{#1}}\xspace}}
\tsc{WGM}
\tsc{QE}
\begin{document}
\let\WriteBookmarks\relax
\def\floatpagepagefraction{1}
\def\textpagefraction{.001}

% Short title
\shorttitle{Muldoon et al.}    

% Short author
\shortauthors{Muldoon et al.}  

% Main title of the paper
\title [mode = title]{Mining Explainable Predictive Features for Water Quality Management}  

% Title footnote mark
% eg: \tnotemark[1]
\tnotemark[1] 

% Title footnote 1.
% eg: \tnotetext[1]{Title footnote text}
\tnotetext[1]{This research was funded, in part, by the European Regional Development Fund.} 

% First author
%
% Options: Use if required
% eg: \author[1,3]{Author Name}[type=editor,
%       style=chinese,
%       auid=000,
%       bioid=1,
%       prefix=Sir,
%       orcid=0000-0000-0000-0000,
%       facebook=<facebook id>,
%       twitter=<twitter id>,
%       linkedin=<linkedin id>,
%       gplus=<gplus id>]

\author[1]{Conor Muldoon}[orcid=0000-0003-1381-2561]

% Corresponding author indication
\cormark[1]

% Footnote of the first author
%\fnmark[1]

% Email id of the first author
\ead{conor.muldoon@ucd.ie}

% URL of the first author
\ead[url]{https://people.ucd.ie/conor.muldoon}

% Credit authorship
% eg: \credit{Conceptualization of this study, Methodology, Software}
\credit{Conceptualization of the article, Methodology, Software, Writing}

\author[1]{Levent G\"{o}rg\"{u}}[orcid=0000-0001-7566-3421]

% Footnote of the second author
%\fnmark[2]

% Email id of the second author
\ead{levent.gorgu@ucd.ie}

% URL of the second author
\ead[url]{https://people.ucd.ie/levent.gorgu}

% Credit authorship
\credit{Software}

\author[2]{John J. O’Sullivan}[orcid=0000-0002-9524-1274]

% Footnote of the second author
%\fnmark[2]

% Email id of the second author
\ead{jj.osullivan@ucd.ie}

% URL of the second author
\ead[url]{https://people.ucd.ie/jj.osullivan}

% Credit authorship
\credit{Funding acquisition-Supporting}

\author[3]{Wim G. Meijer}[orcid=0000-0003-1302-4338]

% Footnote of the second author
%\fnmark[2]

% Email id of the second author
\ead{wim.meijer@ucd.ie}

% URL of the second author
\ead[url]{https://people.ucd.ie/wim.meijer}

% Credit authorship
\credit{Funding acquisition-Equal}

\author[4]{Gregory M. P. O’Hare}[orcid=0000-0002-5124-1686]

% Footnote of the second author
%\fnmark[2]

% Email id of the second author
\ead{gregory.ohare@tcd.ie}

% URL of the second author
\ead[url]{https://www.scss.tcd.ie/personnel/ohareg}

% Credit authorship
\credit{Funding acquisition-Supporting, Review and editing-Supporting}

%John J. O’Sullivan
%Wim Meijer
%Gregory M. P. O’Hare

% Address/affiliation

\affiliation[1]{organization={UCD School of Computer Science},
            addressline={University College Dublin}, 
            city={Dublin},
            country={Ireland}}
            
\affiliation[2]{organization={UCD School of Civil Engineering},
            addressline={University College Dublin}, 
            city={Dublin},
            country={Ireland}}

            % Address/affiliation
\affiliation[3]{organization={UCD School of Biomolecular and Biomedical Science, UCD Earth Institute, UCD Conway Institute},
            addressline={University College Dublin}, 
            city={Dublin},
            country={Ireland}}

% Address/affiliation
\affiliation[4]{organization={School of Computer Science and Statistics, O'Reilly Institute},
            addressline={Trinity College Dublin}, 
            city={Dublin},
			country={Ireland}}

% Corresponding author text
\cortext[1]{Corresponding author}

% Footnote text
%\fntext[1]{Foot foot I have a foot}

% For a title note without a number/mark
%\nonumnote{}

% Here goes the abstract
\begin{abstract}
With water quality management processes, identifying and interpreting relationships between features, such as location and weather variable tuples, and water quality variables, such as levels of bacteria, is key to gaining insights and identifying areas where interventions should be made. There is a need for a search process to identify the locations and types of phenomena that are influencing water quality and a need to explain how the quality is being affected and which factors are most relevant. This paper addresses both of these issues. A process is developed for collecting data for features that represent a variety of variables over a spatial region and which are used for training models and inference. An analysis of the performance of the features is undertaken using the models and Shapley values. Shapley values originated in cooperative game theory and can be used to aid in the interpretation of machine learning results. Evaluations are performed using several machine learning algorithms and water quality data from the Dublin Grand Canal basin.  
\end{abstract}

% Use if graphical abstract is present
%\begin{graphicalabstract}
%\includegraphics{}
%\end{graphicalabstract}

% Research highlights
\begin{highlights}
\item An algorithm for selecting features of location, variable tuples is developed for predicting water quality data and identifying areas for potential interventions.
\item The selected features are analysed using Shapley values to aid with explainability.
\item A novel process is introduced to obtain weather forecast data over a spatial and temporal extent.
\item An evaluation of the methods is performed using real world data from the Dublin Grand Canal basin, Dublin City Council, and Met \'{E}ireann, Ireland's meteorological service.
\end{highlights}

% Keywords
% Each keyword is seperated by \sep

\maketitle

% Main text
\section{Introduction}\label{int}

Process mining is a family of techniques that support the analysis of operational processes, in terms of key performance indicators, using event data \cite{van2012process}. Process mining can be used in number of ways, such as in identifying insights into current processes or in identifying actions or places within workflows where interventions should be made to improve performance. Although processing mining is typically used in the context of commercial business environments, there is crossover to other areas where processes play an important role, such as in water quality management processes administered by local government authorities or citizen science projects that use the Business Process Model and Notation (BPMN) \cite{higgins2016citizen}. In the case of water quality management, traditional event log data from information technology systems is often lacking in that many tasks, such as the manual sampling of water and the microbial culturing by biologists and laboratory technicians to identify faecal coliforms, are not performed using computers and are not logged. Nevertheless, it is likely that techniques developed to aid explainability and in the evaluation of machine learning algorithms in such cases will prove using in traditional process mining systems where similar problems must be addressed. This paper focuses on mining suitable features to perform inference for the level of bacteria, and specifically Enterococci and Escherichia coli (E. coli), in water and on aiding in explaining the inferences made. Thus, improving the water quality in term of levels of Enterococci and E. coli can be viewed as the key performance indicators.

Several approaches have been developed in process mining that make use of Shapley values \cite{shapley1953quota} for interpretability \cite{stevens2022quantifying,pishgar2022process,galanti2020explainable} and in the interpretability of water quality \cite{mahjouri2010game,park2022prediction}. This paper differs from prior research in this area in a number of ways. The application of this work is not with regard to the output of water treatment plants, rather it is in quantifying water quality within an urban environment that hitherto has unknown or unquantified pollutant sources other than those related to treatment plants. A search process is developed that aids in identifying sources and analysis is performed using alternative algorithms. Furthermore, a novel process for collecting data from meteorological services is adopted using Unison \cite{muldoon2021a,muldoon2021b}, which is an open-source system developed by the authors.

There are a number of factors that influence the number of faecal coliforms in water, such as leaking sewage pipes, sewage overflows, agricultural slurry that washes off into streams, and the weather, particularly precipitation. High levels of precipitation often exacerbates the problem of poor water quality, but it can help to identify where the pollution is coming from if precipitation levels are monitored over a wide spatial extent. In Section \ref{uni}, the paper discusses Unison, which enables weather forecast data from Met \'{E}ireann, Ireland's meteorological service, to be tracked over a wide region at a large number of points. The paper details a search process that enables features for machine learning to be identified over the spatial region and over several weather variables, such as precipitation, global radiation, dew point, values etc. Several machine learning algorithms are evaluated in terms of performing inference for the levels of bacteria using this data. Furthermore, the features are analysed in term of their Shapley values to aid in explaining the performance of the models and to identify the most important features for water quality.

The objectives of this research are the following:
\begin{enumerate}
\item Select the set of features of location, variable tuples that optimise the water quality predictive accuracy in terms of Enterococci and E. coli.
\item Evaluate the efficacy of alternative machine learning methods.
\item Evaluate the accuracy of forecast precipitation data.
\item Analyse the selected features for training the models using Shapley values to aid in explaining predicted water quality levels.
\end{enumerate}

\section{Material and methods}\label{mat}
This section discusses the processes, technologies, and methods used in this research. Sections \ref{data} covers how the data was obtained. Section \ref{learn} discusses the machine learning algorithms evaluated. In Section \ref{shap}, Shapley values are discussed, which are used to help explain the machine learning inferences in the evaluation. The search process for selecting features is discussed in Section \ref{search}.

\subsection{Data Acquisition}
\label{data}
The data used in the evaluation includes weather forecast data, which was obtained via the Unison system (Section \ref{uni}), the results of microbial culturing of water samples (Section \ref{wats}), and water gauge data for rainfall (Section \ref{watg}). 

\subsubsection{Unison}
\label{uni}

Unison is an open-source system \cite{muldoon2021b,muldoon2021a} developed by the authors that enables the tracking, visualisation, and storage of weather forecast data from meteorological services. Specifically, meteorological services that use the HARMONIE-AROME model and schema \cite{bengtsson2017harmonie}. HARMONIE-AROME is used by the meteorological services of 10 countries within Europe, including Met \'{E}ireann, Ireland's meteorological service. One of the reasons the system was developed is that HARMONIE-AROME endpoints, which make forecast data available, don't provide historical data. In this research, we wish to use such data to train machine learning models. 

At a high-level, Unison works by periodically connecting to a HARMONIE-AROME endpoint for each location being tracked, downloading and parsing the data, storing and indexing it in a database, and enabling the data to be queried remotely via an API. It also provides functionality to enable users to add or remove locations, etc. along with the associated weather data. The Met \'{E}ireann HARMONIE-AROME model is updated every 6 hours, so Unison connects to the system four times per day and overwrites prior forecast data with outputs from the updated model. The forecast granularity for each weather variable is one hour.

For the deployment of Unison used in this work, close to three hundred locations were tracked and the weather variables included cloudiness, cloud level (low, medium, and high), dew point, global radiation, humidity, precipitation (median, 20\textsuperscript{th} percentile, and 80\textsuperscript{th} percentile), pressure, temperature, wind speed, and wind direction.

\subsubsection{Water Samples}
\label{wats}

Water samples were taken manually by Dublin City Council (DCC) from January 2021 to the end of May 2022 from seven sites in the Dublin Grand Canal basin. Assessment of water quality, in this paper, involves the analysis of two types of bacteria, namely E. coli and Enterococci. These bacteria and the number of coliforms represent indicators of water quality. The number of coliforms and the bacteria levels were determined by microbial culturing, which was performed by biologists in the Dublin City Central Laboratory. The sample data included the levels of ammonia, chemical oxygen demand, conductivity, dissolved oxygen, nitrate, acidity, phosphorus, suspended solids, temperature, total oxidised nitrogen, and total biological oxygen.

\subsubsection{Water Guages}
\label{watg}
To evaluate the accuracy of the forecast data for precipitation, data was obtained from eight DCC water gauges for rainfall. The gauge data was taken to be the ground truth for the gauge locations. A system was developed that operates in a somewhat simplified manner to Unison (Section \ref{uni}). The system periodically downloads the data from DCC's, stores and indexes it, and makes it available via an API. The granularity of the gauge data was fifteen minutes, so it was aggregated to be an hour to be consistent with the forecast data.

\subsection{Learning Algorithms and Technologies}
\label{learn}
Four supervised machine learning algorithms for regression are evaluated in Section \ref{res}. The algorithms can be grouped in a number of ways. Two kennel methods were used, namely the Support Vector Regression (SVR) and Gaussian process regression. The implementation of the Gaussian process and neural network were GPU accelerated. Specifically, the NVIDIA Tesla M10 was used. Gaussian processes were the only Bayesian approach adopted.

\subsubsection{Linear Regression}

Linear regression can be viewed as a basic form of machine learning. The Python Scikit-learn machine learning package was used to perform ordinary least squares linear regression. The Scikit-learn implementation can be used to calculate the coefficient of determination, or R-squared value, in addition to the predicted values. The R-squared value is the residual sum of squares and can be used as an indictor of how well the model fits the data.

One of the main advantages of linear regression is its simplicity, which aids in interpretability. The primary disadvantage, however, is that it not suitable in cases where there are nonlinearities in the mapping between the independent and dependent variables.

\subsubsection{Support Vector Regression}
\label{svr}

In addition to linear regression, the Scikit-learn package was used for the SVR implementation. SVRs are Kernel methods, which were the dominant form of machine learning algorithm prior to the more recent success of neural networks. SVRs are a variant of Support Vector Machines (SVMs) \cite{drucker1996support} that solve regression problems rather than classification. They are robust prediction models and are based on Vapnik–Chervonenkis theory \cite{vapnik1999nature}. SVRs, and kernel methods for regression in general, enable non-linear regression using what's known as the kernel trick, whereby inputs are implicitly mapped to high-dimensional feature spaces. The default kernel, namely the radial basis function, was used for the evaluations.

SVRs are effective in high dimensional cases and in cases where the number of dimensions is greater than the number of samples. They are memory efficient in that they use a subset of training points in the decision function or support vectors. Additionally, they are versatile in that different kernel functions can be specified for the decision function. A disadvantage to SVRs, however, is that they are prone to overfitting in cases where the number of features is much greater than the number of samples.

\subsubsection{Neural Networks and PyTorch}

In situations where there is a large amount of data and significant computational resources available, neural networks and, in particular deep neural networks with many hidden layers, have achieved significant empirical success for many applications. PyTorch has become one of the most popular neural network frameworks and enables scalable distributed training and performance along with GPU acceleration. The library contains many useful features that ease the programming burden including automatic differentiation \cite{van2018automatic}. A couple of disadvantages with deep neural networks, though, is that they tend to overfit if there is little data available and they can be more difficult to interpret than alternative approaches.

The neural network evaluated in Section \ref{res} has been implemented in PyTorch. The network has two hidden layers of width ten and Rectified Linear Unit (ReLU)  activation units to introduce non-linearity. The weights of the hidden layers were initialised using the Xavier uniform distribution and the biases were initialised to zero. A mean squared error loss function and the Adam optimizer \cite{kingma2014adam} were used for training. 

\subsubsection{Gaussian Processes and GPyTorch}
\label{gp}

Despite their great empirical success, much work must still be done to put feedforward neural networks on a sound theoretical footing. Under broad conditions, as the width of neural network architectures increase, the implied random function converges in distribution to a Gaussian process \cite{neal2012bayesian,matthews2018gaussian}.

Gaussian processes \cite{rasmussen2003gaussian} are kernel methods and, as with SVRs, make use of the kernel trick (Section \ref{svr}). Gaussian processes differ from SVRs, however, in that they are Bayesian methods. One of the advantages of Gaussian processes for regression over SVRs is that they predict distributions rather than point values and the variance of the predicted distribution can be viewed as a measure of the uncertainty of the predicted value or mean. A disadvantage of Bayesian methods, however, is that they tend to be more computationally expensive than alternative frequentist approaches. 

The Gaussian process model used in the evaluation was implemented using GPyTorch \cite{gardner2018gpytorch} and a radial basis function kernel. GPyTorch enables the efficient and modular implementation of Gaussian processes. It is implemented in PyTorch and enables GPU acceleration. For training, the marginal log likelihood was maximised using the Adam optimizer \cite{kingma2014adam}. The marginal likelihood is the model evidence or the probability of the data given the model.

\subsection{Shapley Values}
\label{shap}

Shapley values originate in cooperative game theory and provide a principled way to explain inferences in machine learning by conceptualising a machine learning model trained on a set of features as a value function on a coalition of agents. Shapley values enable the value contributed by a feature to a prediction to be calculated.

For a set $N$ of $n$ agents and a characteristic function that maps subsets of agents to real numbers $v \colon 2^N \to \mathbb{R}, v ( \emptyset ) = 0$ and a set $S$ that is a coalition of agents, the Shapley value $\varphi_i(v)$ for an agent $i$ is defined as follows:
\begin{equation}\label{shapley}
\varphi_i(v)=\sum_{S \subseteq N \setminus
\{i\}} \frac{|S|!\; (n-|S|-1)!}{n!}(v(S\cup\{i\})-v(S))
\end{equation}
Consider a coalition being built up one agent at a time. The marginal increase in the utility or the contribution of adding an agent $i$ to the coalition is $v(S\cup\{i\})-v(S)$. The Shapley value represents the average contribution over all possible permutations in which the coalition is formed. An alternative, more intuitive, expression of the Shapley value, but which is not used in practice, is as follows:
\begin{equation}
\varphi_i(v) = \frac{1}{n} \sum_{S \subseteq N \setminus \{i\}} \binom{n-1}{|S|}^{-1} (v(S \cup \{i\}) - v(S))
\end{equation}

One of the main drawbacks of computing Shapley values is that it is computationally intractable (NP hard) for a large number of agents. Under certain conditions on the characteristic function, such as it being submodular \cite{liben2012computing}, polynomial-time approximation schemes can be used. In general, however, there is no guarantee on the performance of approximate or greedy schemes. When there are only a small number of agents or features, this is not an issue.

In this research, SHapley Additive exPlanation (SHAP) values were used and were estimated using the SHAP framework \cite{lundberg2017unified}. The SHAP framework is available as a Python package. The term  SHAP value is more specific than Shapley value and refers to Shapley values that are applied to a conditional expectation function of a machine learning model. 

\subsection{Search Process}
\label{search}

This section discusses the process for selecting location and weather variable tuples as features for the machine learning algorithms evaluated in Section \ref{res}. Prior to the search process operation, weather forecast data for points over the spatial region of interest is gathered using Unison (Section \ref{uni}) along with water samples, which are processed in the laboratory to determine bacteria levels (Section \ref{wats}). The meteorological service does not provide historical data, thus, Unison must be in operation over the entire period that water sampling occurs. Once laboratory results are available and the data for Unison has been obtained for the time period, the selection process begins.

Two metrics are used in determining the locations and weather variables to select. Pearson's correlation coefficient $r$ and Kendall's rank correlation coefficient $\tau$. For a pair of random variables $(X,Y)$ where $\sigma_X$ is the standard deviation of $X$, $\sigma_Y$ is the standard deviation of $Y$, $\mu_X$ is the mean of $X$, and $\mu_Y$ is the mean of $Y$, Pearson's $r$, $\rho_{X,Y}$, can be defined as follows:

\begin{equation}
\rho_{X,Y} = \frac{\mathbb{E}[(X-\mu_X)(Y-\mu_Y)]}{\sigma_X\sigma_Y}
\end{equation}

\begin{algorithm}
\caption{Search process feature selection}\label{alg:cap}
\begin{algorithmic}[1]
\State $map \gets Map()$
\For{each GCB site as $s$}
	\For{Enterococci and E. coli as $wqv$}
		\State $qual \gets loadQualityData(s, wqv)$
		\For{each location being tracked as $l$}
			\For{each weather variable as $wv$}
				\State $weather \gets loadWeatherData(l, wv)$
				\State $agg \gets []$
				\For{each sample in $qual$ as $samp$}
					\State $a=aggregate(weather, samp['date'])$
					\State $agg.append(a)$
				\EndFor
				\State $r, rp \gets pearson(qual, agg)$
				\State updateBest(rp, s, wv, wq, 'pearson', map, r, l)
				\State $\tau, tp \gets kendall(qual, agg)$
				\State updateBest(tp, s, wv, wq, 'kendall', map, $\tau$, l)
			\EndFor
		\EndFor
    \EndFor
\EndFor
\Procedure{updateBest}{p, s, wv, wqv, ct, map, c, l}
	\If{p < 0.05}
		\State $key \gets s + wv + wqv + ct$
		\State $current \gets map.get(key)$
		\If {current == NONE or abs(c) > abs(current['correlation'])}
			\State map.put(key,\{'p-value': p, 'correlation': c, 'location': l\})		
		\EndIf
	\EndIf
\EndProcedure
\end{algorithmic}
\end{algorithm}

One of the problems with Pearson's $r$ is that it won't necessarily identify non-linear relationships between independent and dependent variables. Kendall's $\tau$ caputures monotonic non-linearities and it is less vulnerable to outliers than Pearson's r, but is less powerful in that it is a non-parameteric correlation and less information is used in the calculation. Pearson's $r$ uses information about the mean and deviation from the mean, but non-parametric correlations only use ordinal information. Equation \ref{ken} defines Kendall's $\tau$ for of a set of $n$ observations ${\displaystyle (x_{1},y_{1}),...,(x_{n},y_{n})}$ of X and Y.

\begin{equation}\label{ken}
\tau = \frac{2}{n(n-1)}\sum_{i<j} sign(x_i-x_j)sign(y_i-y_j)
\end{equation}

Algorithm \ref{alg:cap} gives the pseudocode for the search process feature selection. Initially, a map is created that is used to store the best locations and weather variables for each site in the Grand Canal basin and for both Enterococci and E. coli. For each site and  water quality variable (Enterococci and E. Coli), the process loops through all locations where weather data is being recorded and all weather variables (Section \ref{uni}) and calculates the Pearson $r$ and Kendall $\tau$. If the absolute values of the correlations are greater than those currently stored for the keys, the map is updated. The key for the map is a concatenation of the site, the weather forecast variable, the water quality variable, and the correlation type (Pearson or Kendall). The location that is stored is the location where the weather forecast data was recorded for, not the site location. On Line 10, prior to computing the correlations, the weather data, which is hourly, is aggregated into a window for the date each water sample was taken.

\section{Results}\label{res}
In Section \ref{comp}, a comparison of the accuracy of the inferences of the machine learning models is given along with a sample of the output of the search process. Section \ref{shapa} gives an analysis for multiple regression using Shapley values. An evaluation of the accuracy of forecast precipitation levels is provided in Section \ref{evalp}. For the results in Sections \ref{comp} and \ref{shapa}, where the machine learning  models were used, 80\% of the sample data was used for training and 20\% for testing. For the water quality data, the laboratory results, in terms of the numbers of Enterococci and E. coli coliforms, from 385 water samples were used.

\subsection{Comparison of Models}
\label{comp}

\begin{table}[]
\caption{Sample of selected features}\label{sftab}
\begin{tabular}{lllllll}
\toprule
Forecast location&Site& Weather variable&Water quality varialble& Correlation type&Correlation&P-value\\
\midrule
Dublin 120905 & 3 & Enterococci & Precipitation (80\textsuperscript{th})  & Pearson & 0.271 & 0.023 \\
Dublin 123559 & 1 & E. coli &  Precipitation (20\textsuperscript{th}) & Pearson  & -0.32 & 0.043 \\
Dublin 121968 & 2 &  E. coli  & Global radiation & Kendall & 0.281 & 0.036 \\
Dublin 120384 & 5 & Enterococci & Dew point & Pearson & 0.336 & 0.029 \\
Dublin 120905 & 6 & E. coli & Precipitation (Median) & Kendall &-0.21  & 0.019 \\
Dublin 123559 & 1 & E. coli & Precipitation (80\textsuperscript{th}) & Pearson & -0.31 & 0.047\\
Dublin 120905 & 7 & E. coli & Precipitation (80\textsuperscript{th}) & Kendall & -0.20 & 0.011 \\ 
Dublin 124618 & 1 & E. coli & Temperature & Kendall & 0.356 & 0.001\\
\bottomrule
\end{tabular}
\end{table}

The search process in Section \ref{search} produces a large amount of information regarding the weather forecast locations and weather forecast variables most correlated with site locations. Table \ref{sftab}\footnote{The numbers in the forecast location names in tables \ref{sftab} and \ref{modelcomp} are GRIB indices and are consistent with the indicies used in the M{\'E}RA \cite{gleeson2017met} meteorological climate reanalysis dataset.} provides a representative sample of this information. Unison (Section \ref{uni}) tracks data for large number of points. The search process determines which points will be used in training.   

The predictions of the models were compared using the Root-Mean-Square Error (RSME). For an estimator $\hat{\theta}$ and an estimated parameter $\theta$, the RMSE is defined as follows: 

\begin{equation}
\operatorname{RMSE}(\hat{\theta}) = \sqrt{\mathbb{E}((\hat{\theta}-\theta)^2)}
\end{equation}

The results for the comparison of the models in terms of the RMSEs are given in Table \ref{modelcomp}. The Gaussian process model performs the best for three sites, the SVR model for two sites, and linear and neural network models for one site each. A large number of forecast locations are identified by the search process. Thus, only the results for the forecast locations where the RMSE of one of the models was the lowest overall are shown.  

\begin{table}[]
\caption{Comparison of models by site for Enterococci}\label{modelcomp}
\begin{tabular}{lllllll}
\toprule
Site & Forecast location & Linear RMSE & SVR RMSE & GP RMSE &  NN RMSE\\
\midrule
\\
1 & Dublin 123552 & 270.53 & 87.59 & 68.88 &193.93  \\ 
2 & Dublin 121968 & 912.83 & 181.44 & 146.66 & 244.34\\
3 & Dublin 123030 & 1164.53 & 1029.79 & 1047.19 & 907.46\\
4 & Dublin 121966 & 102.03 & 93.02 & 84.01 & 117.25\\
5 & Dublin 124618 & 193.79 & 162.93 & 183.26 & 225.25\\
6 & Dublin 120905 & 385.43  & 407.66 & 423.18 & 397.31 \\
7 & Dublin 121966 & 153.97 & 141.65 & 159.17 & 149.08 \\
\bottomrule
\end{tabular}
\end{table}

For the results in Table \ref{modelcomp}, the independent variables for weather were normalised, but the Enterococci data was not. The models, and in particular, the kernel-based models performed quite poorly on raw data for the E. coli. This was due to the E. coli data spanning several orders of magnitude. As such, results for log normalised E. coli and Enterococci are given in tables \ref{lnecoli} and \ref{lnent} respectively. For the log normalised E. coli results, the SVM model performed best for four sites. The Gaussian process, linear, and neural networks models were best for one site each. For the log normalised Enterococci results the linear model performed best for four sites, the neural network of two sites, and the Gaussian process for one site.

\begin{table}[]
\caption{Comparison of models by site for log normalised E. coli}\label{lnecoli}
\begin{tabular}{lllllll}
\toprule
Site & Forecast location & Linear RMSE & SVR RMSE & GP RMSE &  NN RMSE\\
\midrule
\\
1 & Dublin 124618 & 0.192 & 0.156 & 0.164 & 0.215 \\
2 & Dublin 120907 & 0.257 & 0.317 & 0.275 & 0.284 \\
3 & Dublin 120907 & 0.346 & 0.268 & 0.283 & 0.309 \\
4 & Dublin 121966 & 0.290 & 0.291 & 0.288 & 0.208 \\
5 & Dublin 124618 & 0.202 & 0.111 & 0.144 & 0.243 \\
6 & Dublin 121966 & 0.275 & 0.259 & 0.251 & 0.646 \\
7 & Dublin 120904 & 0.252 & 0.203 & 0.218 & 0.597 \\
\bottomrule
\end{tabular}
\end{table}

\begin{table}[]
\caption{Comparison of models by site for log normalised Enterococci}\label{lnent}
\begin{tabular}{lllllll}
\toprule
Site & Forecast location & Linear RMSE & SVR RMSE & GP RMSE &  NN RMSE\\
\midrule
\\
1 & Dublin 120903 & 0.325 & 0.446 & 0.478 & 0.483 \\
2 & Dublin 121968 & 0.635 & 0.735 & 0.770 & 0.875 \\
3 & Dublin 120905 & 0.776 & 0.879 & 0.839 & 1.038 \\
4 & Dublin 121966 & 0.690 & 0.838 & 0.779 & 0.684 \\
5 & Dublin 124618 & 0.763 & 0.666 & 0.587 & 0.516 \\
6 & Dublin 121966 & 0.946 & 0.924 & 0.905 & 0.929 \\
7 & Dublin 120905 & 0.572 & 0.717 & 0.702 & 0.594 \\
\bottomrule
\end{tabular}
\end{table}

\subsection{Shapley Value Analysis}
\label{shapa}

\begin{figure}
\begin{center}
  \includegraphics[width=12.5cm]{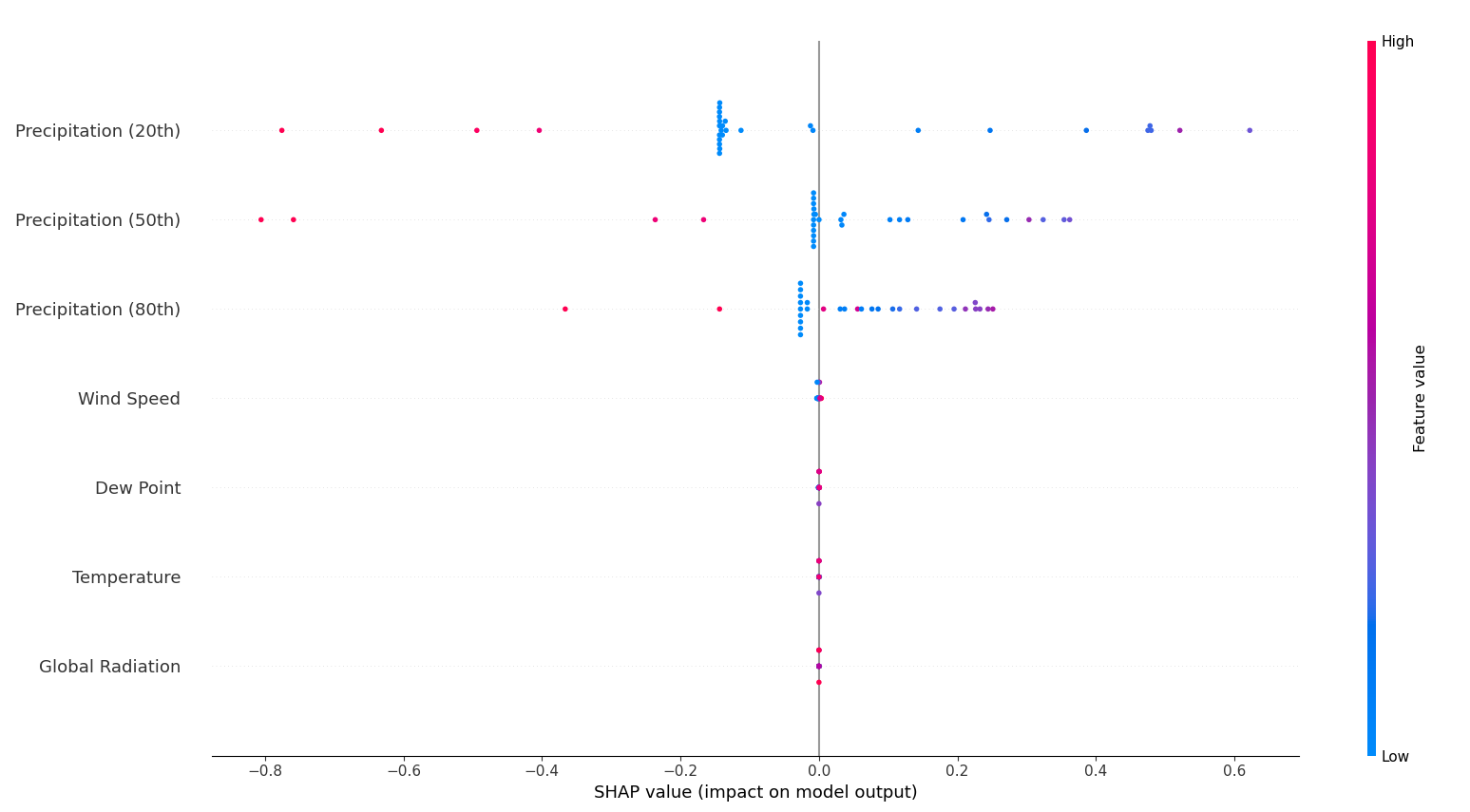}
  \caption{SVR beeswarm chart for Enterococci at Site 4.}
   \end{center}
  \label{fig:beeswarm}
\end{figure}

Figure \ref{fig:beeswarm}, shows a beeswarm chart for the SHAP values (see Section \ref{shap}) of a SVR model for Enterococci at Site 4. Beeswarm charts provide an information-dense view how of features in a dataset impact the model’s output. The colour of the points indicates the value\footnote{Value here refers to the observed value, such as the forecast precipitation level, rather than the degree to which the feature is valuable to the output.} of the features, whereas the x-coordinate is determined by the SHAP value of the feature. Features can pile-up on feature rows to indicate density. Features are displayed in term of largest impact on the model from top to bottom. Figure \ref{fig:beeswarm} indicates that the precipitation variables have the greatest impact on the model output. This is also broadly true for other sites and when using other types of model. One thing to note on the chart is that SHAP values can be negative. Negative SHAP values can be interpreted as the feature having a negative impact on the model output or prediction. Not all weather forecast variables are included in the chart in that not all the variables had correlations with bacterial levels at the site and were not used in training the model. 

\begin{figure}
\begin{center}
  \includegraphics[width=12.5cm]{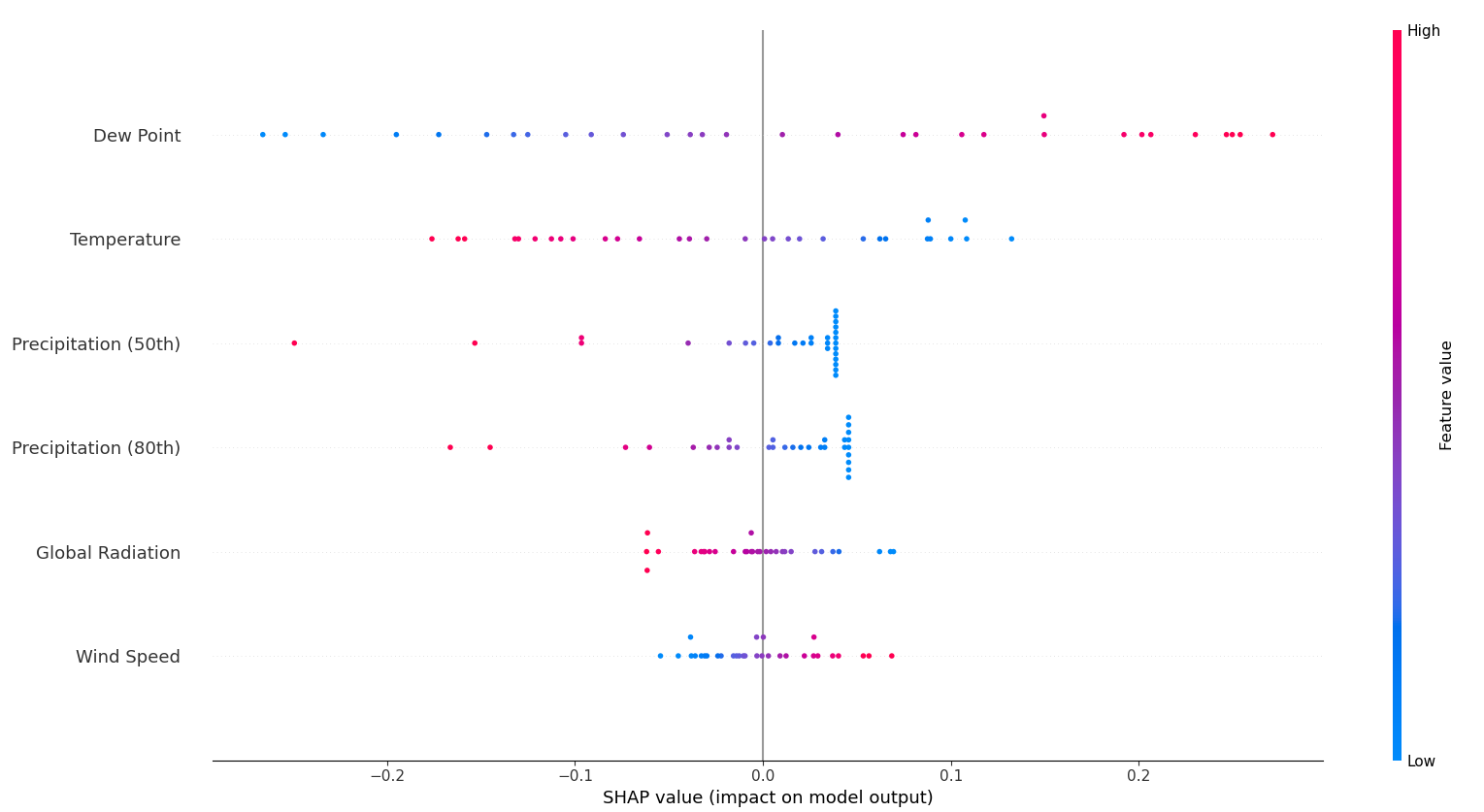}
  \caption{Linear beeswarm chart for E. coli at Site 1.}
 \end{center}
  \label{fig:beeswarm2}
\end{figure}

\begin{figure}
\begin{center}
  \includegraphics[width=12.5cm]{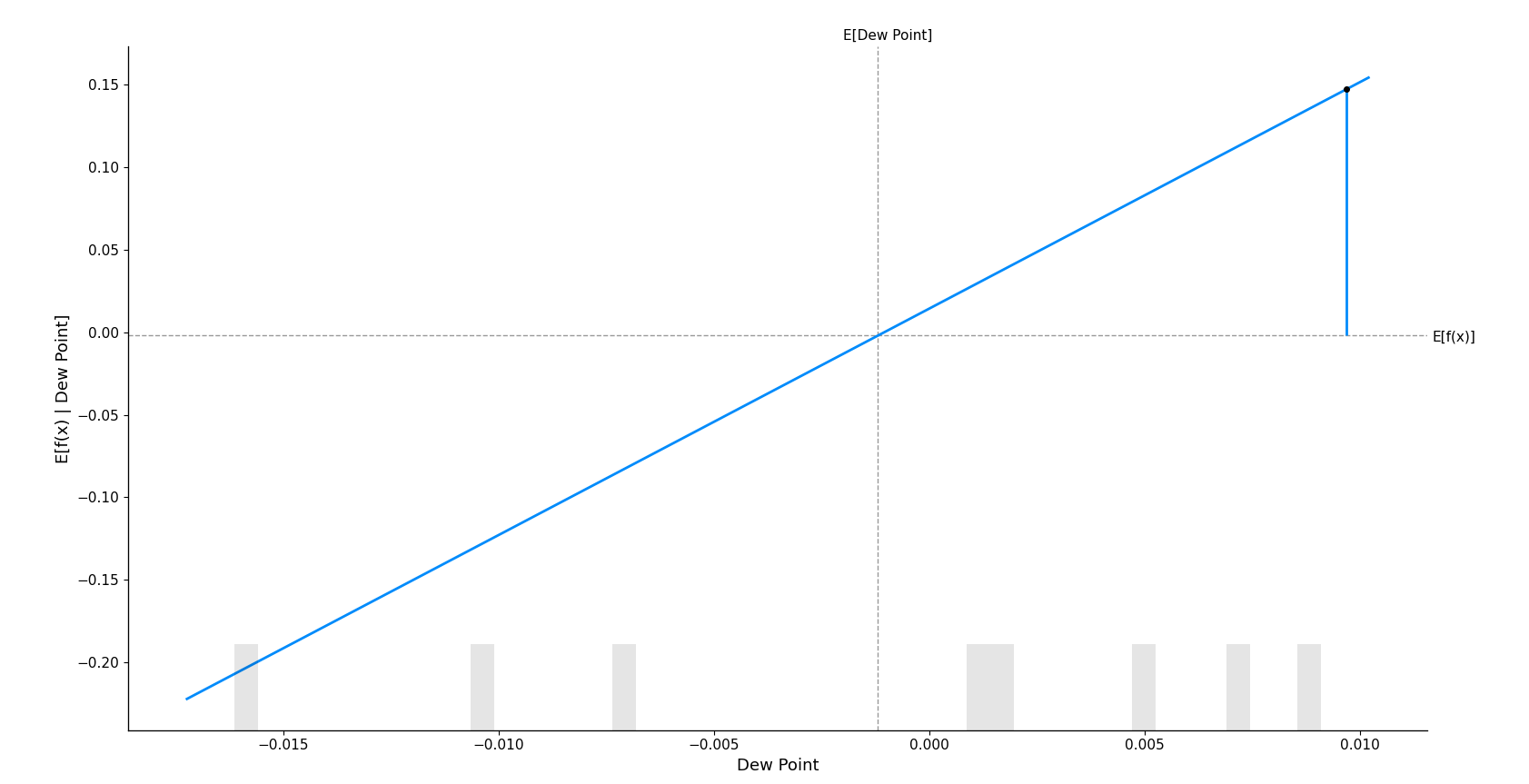}
  \caption{Partial dependence graph for dew point data for a linear model at Site 1.}
 \end{center}
  \label{fig:dep}
\end{figure}

\begin{figure}
\begin{center}
  \includegraphics[width=12.5cm]{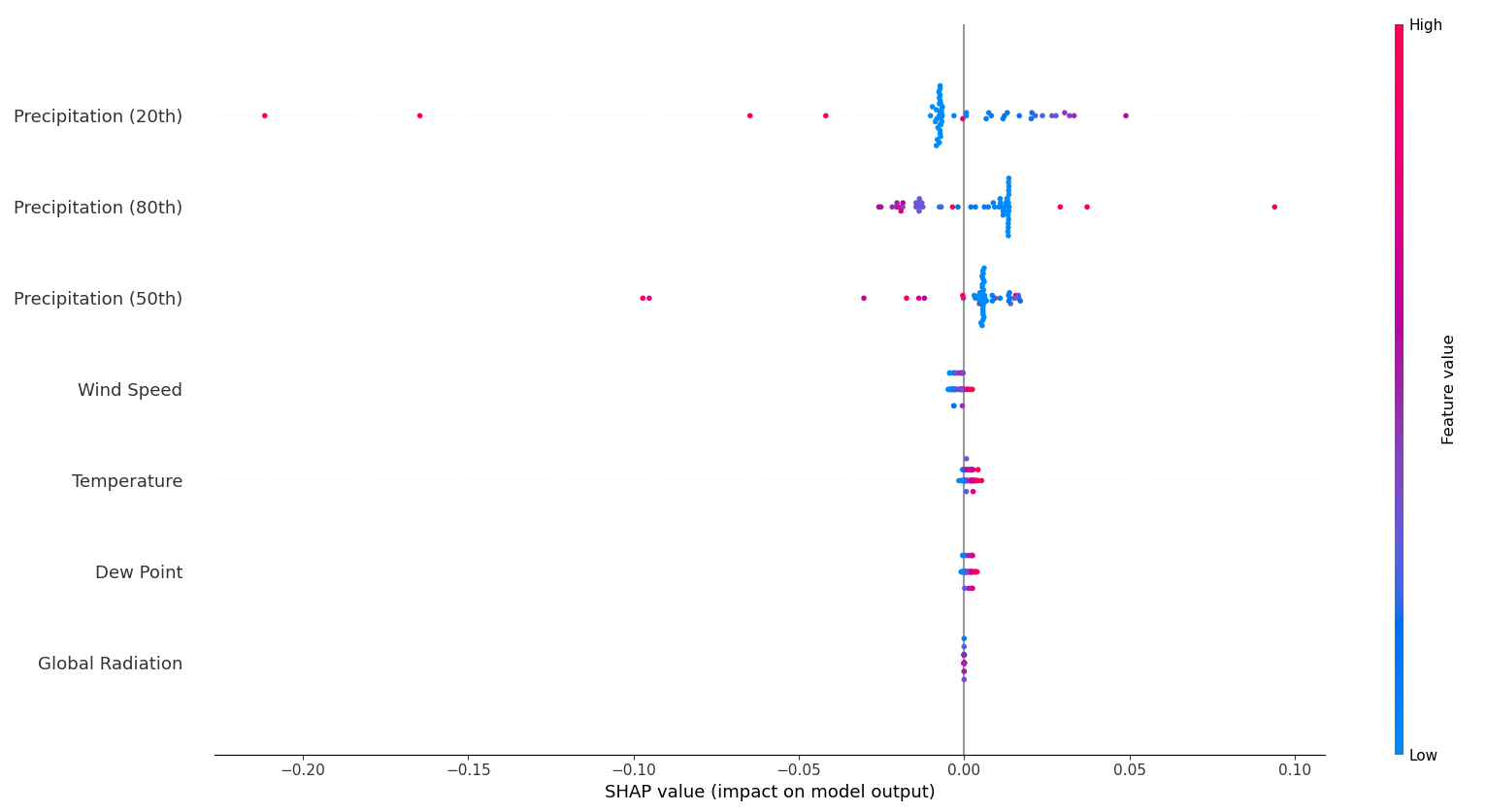}
  \caption{SVR beeswarm chart for E. coli at Site 3.}
 \end{center}
  \label{fig:beeswarm3}
\end{figure}

Figure \ref{fig:beeswarm2} illustrates a beeswarm chart for a linear model for predicting E. coli at Site 1. In this case, the forecast dew point and temperature have a greater impact on the model output than precipitation. In general, different sites, different forecast locations, different bacterium types, and different model types will produce different beeswarm charts and variations on the order of impact of features along with different distributions of SHAP values per feature. Nevertheless, there are general trends regarding the most impactful features. Figure \ref{fig:beeswarm3} illustrates the beeswarm chart for the SVR model for predicting E. coli at Site \ref{fig:beeswarm3}. Although SHAP values, in general, are NP hard to compute, in the case of linear models, they are simple and can be read from partial dependence graphs, such as Figure \ref{fig:dep}.

\subsection{Evaluation of Forecast Precipitation}
\label{evalp}
Given the importance precipitation data plays in model inference (see Section \ref{shapa}), an evaluation was performed to determine the accuracy of foceast precipitation. DCC rain gauge data and forecast data was obtained from October 2021 to September 2022. The gauge data was taken as the ground truth and the Time Lagged Cross Correlation (TLCC) method was used to determine if there was there a delay between the forecast precipitation level and observed water gauge level. The results in Table \ref{raineval} are given for the delay rather than the offset correlation. A very large number of samples (over 6000) were used, so the p-value was close to zero in all cases. For each gauge, the closest spatial location where weather data was recorded was determined and used in the evaluation. Table \ref{raineval} gives the results for the forecast 20\textsuperscript{th} percentile. Similar results were received for the median and 80\textsuperscript{th} percentile. The TLCC method identifies that in all cases the forecast either leads the values by an hour\footnote{The forecast data granularity is hourly.}(lag of minus -1) or are in synchronicity in terms of their highest of correlation values.

%Gauge location, forecast location, 20th r, 20th lag, median r, median lg, 80th r, 80th lag

\begin{table}[]
\caption{Precipitation (20\textsuperscript{th} percentile) forecast evaluation}\label{raineval}
\begin{tabular}{lllllllll}
\toprule
Gauge  & Forecast location &  20\textsuperscript{th} r & 20\textsuperscript{th} lag\\
\midrule
\\
Ballymun & Ballymun Library & 0.59 & -1 \\ 
Dunboyne & Dunboyne & 0.56  & 0\\
West Pier &  West Pier & 0.63& -1 \\
Enniskerry & Shanganagh & 0.67 & -1\\
Kippure & Liffey 6 &  0.65 & -1\\
Arklow &  Liffey 6 & 0.58 & -1\\
Crumlin & Crumlin & 0.64 & 0\\
The Grange PS& Dublin 124084 & 0.55 & -1\\
\bottomrule
\end{tabular}
\end{table}

\section{Discussion}\label{dis}

There are several factors that influence that quality of water in urban environments. Factors, such as discharges from water treatment plants, can be accounted for without the need to identify the pollution source. Other factors are more difficult to elicit. In this paper, a search process is discussed whereby correlations between weather forecast data and water quality data are used to identify locations where weather variables are correlated with bacterial levels. In the search process, forecast data is aggregated for the entire day that each water sample was taken for. Future work will investigate using different time periods or window sizes and ranges. An approach such as TLCC (see Section \ref{evalp}) cannot be used for this, however, in that water samples are taken sporadically and the data does not represent a time series. 

Prior to executing the selection algorithm for the search, close to three hundred points in the Dublin region were added to Unison (Section \ref{uni}) and tracked over an extended period of time. During this period, water samples were taken and processed in the laboratory. The output of the search process is used either directly to identify potential pollution sources or indirectly through the inference of machine learning models and Shapley analysis of such. In the case of using the outputs directly, a local authority official, for example, could investigate areas where precipitation data is highly correlated with poor water quality. With high levels of precipitation, agricultural slurry is more likely to end up in streams and bacteria from leaking pipes is more likely to make it to the sites of interest. It should be noted, however, for some of the model outputs it was found that precipitation was negatively correlated with water quality. In such cases, at the forecast locations, which are different from the site locations, if there are high levels of precipitation, poor quality water gets washed away to somewhere other than a site location.

In the comparison of the models in Section \ref{comp}, it was found that Gaussian processes performed best for inferring Enterococci when raw data was used for training, SVRs performed best for Enterococci when using log normalised data, and linear models for E. coli when using log normalised data. With log normalised data, linear models perform better in that the non-linearity of the data is reduced. For the Gaussian processes exact inference was performed and the exact marginal log likelihood was computed. If a very large number of samples were used this would not be tractable, but several approximate methods are available. It is likely that the neural network models were overfitting, but they did perform best in a number of cases. Further experimentation with the hyperparameters of SVRs, Gaussian processes, and neural networks would likely improve the performance.

To a certain extent, the accuracy of models is dependent on the data that they are trained on. In Section \ref{evalp}, the forecast data for the most impactful feature in the models in general, namely precipitation, was evaluated using TLCC and data from DCC water gauges as the ground truth. It was found that although the forecast precipitation was correlated with gauge data and was either in synchronicity or leading by an hour, they were 100\% correlated. Thus, inaccurate forecasts are a source of error. It should be noted that gauges track data for a single point, however, and may not be representative of the broader region.

Traditionally, machine learning has been viewed as a black-box for representation learning and inference and was often criticised for a lack of interpretability. Significant progress has been made in recent years in relation to interpretability and in particular the development of the SHAP framework, which provides a practical approach to estimating Shapley values and represents a unified framework for several pre-existing methods in the area. SHAP values provide fine-grained insight into model performance and can aid officials to identify relevant factors. The beeswarm charts in Section \ref{shapa} show that the most important features vary depending on site locations and variables under consideration along with the distribution of SHAP values per feature. Being able show how models make specific inferences on a per feature basis increases trust in the models. The more information that can be provided on how features impact the key performance indicators, which in this case related to bacterial levels, the more scope there is for improving and optimising operational processes.

It should be noted that although SHAP values aid in the interpretability of the outputs of models, they should not be interpreted as identifying causal factors. To give an example of misleading SHAP values, from a causality perspective, at a couple of sites wind speed values were found to have an impact on the model output. This, however, is likely due to confounding causes. Future work will investigate the use of structural causal models and Pearl's do-calculus \cite{pearl2018book} to aid with counterfactual reasoning and evaluating the impact of potential interventions.

\section{Conculsions}\label{con}

One of the key objectives of effective water quality management processes is in identifying the locations of pollution sources, along with identifying their types, to enable interventions to be made. In this paper, the use of water quality data and a novel method for collecting data from meteorological services for feature selection is discussed. The features can be either used directly by local authority officials or indirectly used though machine learning models and the SHAP value analysis thereof. 

Four types of machine learning architectures, namely linear regression, support vector regression, Gaussian process, and neural network architectures, were evaluated over selected forecast locations and seven sites in the Dublin Grand Canal basin using the laboratory results for Enterococci and E. coli levels from over a year and a half time frame. Under different circumstances, linear regression, support vector regression, and Gaussian process regression perform best. The neural network was likely overfitting, but experimenting with hyperparameters could potentially improve the performance as with the availability of a greater number of samples. 

There are several factors that influence the output of machine learning models and SHAP values provide a principled way to explain the outputs of the models, which is both informative and improves trust and transparency. Future work will investigate the development of a structural causal model to aid in identifying interventions and changes in operational processes to improve both the efficiency and effectiveness of water quality management. Furthermore, the optimisation of the window size and lag for data aggregation in the selection process will be investigated. 

\section*{Acknowledgement}

The authors would like to acknowledge the support of the EU Ireland Wales European Territorial Co-operation (ETC) programme.

% To print the credit authorship contribution details
\printcredits

%% Loading bibliography style file
%\bibliographystyle{model1-num-names}
\bibliographystyle{cas-model2-names}

% Loading bibliography database
\bibliography{refs}

% Biography
\bio{}
Dr. Conor Muldoon is a research scientist in the School
of Computer Science at UCD with research interests in the areas of Sensor Networks, Multi-Agent Systems, Distributed Artificial Intelligence,
and Ubiquitous Computing. He has held two prestigious Irish Research
Council for Science, Engineering, and Technology (IRCSET) fellowships, namely the INSPIRE Marie Curie International Mobility Fellowship and the Embark Fellowship. He holds a Ph.D. in Computer Science and
a B.Sc. (Honours) degree in Computer and Software Engineering.

\endbio

\bio{}
% Here goes the biography details.
Dr. Levent G\"{o}rg\"{u} s a research scientist in the School of
Computer Science at UCD. He holds a Ph.D. in Computer Science, an
M.Sc. in Ubiquitous and Multimedia Systems, and a B.Sc. in Computer
Engineering. His research interests include Wireless Sensor Networks
and Ubiquitous Computing. He was previously an assistant professor
at the Cyprus International University where he taught courses on advanced database management systems, advanced programming, and
visual programming.
\endbio

\bio{}
Born and brought up in Dublin, Prof. John J. O’Sullivan  studied at Trinity
College Dublin from where he graduated with a Civil Engineering degree
in 1993. He completed an M.Sc. at the Queen’s University of Belfast in
1994 and following this started a seven-year period that would involve
working in a research capacity at the University of Ulster on a number of
externally funded projects. He completed his Ph.D. in 1999 after spending
time at the UK Flood Channel Facility in HR Wallingford and also
spent time at the University of the Witwatersrand, South Africa and
at the University of Adelaide, Australia. Present research interests are
concerned with water resources.
\endbio

\bio{}
Prof. Wim G. Meijer is Professor of Microbiology and
Head of the School of Biomolecular and Biomedical Science at UCD.
His research team works in two thematic areas: water quality and
human/animal health. His water themed research focuses on water
quality of bathing waters and rivers, in particular in relation to faecal
contamination, pathogens and antimicrobial resistance, working closely
with colleagues in other disciplines, with local authorities and national
regulatory bodies. Animal health research is mainly focused on the
Rhodococcus equi, a multi-host pathogen infecting phagocytic cells, and
the role of microbiota in uterine health in relation to the development of
endometritis.

\endbio

\bio{}
Prof. Gregory M. P. O’Hare is Professor of Artificial
Intelligence and Head of the School of Computer Science and Statistics
at Trinity College Dublin and a visiting Professor at UCD. Prof. O’Hare
has over 500 refereed publications of which over 100 are in high impact
journals. He has edited 10 books and has a cumulative career research
grant income of circa C82 Million. His research interests are in the
areas of Artificial Intelligence and Multi-Agent Systems (MAS), Ubiquitous Computing and Wireless Sensor Networks. He is an established
Principal Investigator with Science Foundation Ireland, having been one
of the founders of the CLARITY centre (now INSIGHT) and the very
successful CONSUS collaboration with Origin Enterprises in the area of
Digital Agriculture.

\endbio

\end{document}